\documentclass[letterpaper, 10 pt, conference]{ieeeconf}  

\IEEEoverridecommandlockouts                              

\usepackage{cite}
\usepackage{amsmath,amssymb,amsfonts}
\usepackage{algorithmic}
\usepackage{graphicx}
\usepackage{textcomp}
\usepackage{xcolor}
\usepackage{subcaption}
\def\BibTeX{{\rm B\kern-.05em{\sc i\kern-.025em b}\kern-.08em
    T\kern-.1667em\lower.7ex\hbox{E}\kern-.125emX}}

\title{\LARGE \bf
Calibrating the Full Predictive Class Distribution of 3D Object Detectors for Autonomous Driving
}

\author{Cornelius Schröder$^{1}$, Marius-Raphael Schlüter and Markus Lienkamp$^{1}$
\thanks{This research was funded by the Bavarian Research Foundation.}
\thanks{$^{1}$ Institute for Automotive Engineering,}
\thanks{\hspace{0.25cm}Munich Institute of Robotics and Machine Intelligence,}
\thanks{\hspace{2.5mm}School of Engineering and Design,}
\thanks{\hspace{2.5mm}Technical University of Munich}
\thanks{{\tt\small cornelius.schroeder@tum.de}}
\thanks{\copyright 2025 IEEE}
}

\begin{document}

\maketitle
\thispagestyle{empty}
\pagestyle{empty}

\begin{abstract}
In autonomous systems, precise object detection and uncertainty estimation are critical for self-aware and safe operation. This work addresses confidence calibration for the classification task of 3D object detectors. We argue that it is necessary to regard the calibration of the full predictive confidence distribution over all classes and deduce a metric which captures the calibration of dominant and secondary class predictions. We propose two auxiliary regularizing loss terms which introduce either calibration of the dominant prediction or the full prediction vector as a training goal. We evaluate a range of post-hoc and train-time methods for CenterPoint, PillarNet and DSVT-Pillar and find that combining our loss term, which regularizes for calibration of the full class prediction, and isotonic regression lead to the best calibration of CenterPoint and PillarNet with respect to both dominant and secondary class predictions. We further find that DSVT-Pillar can not be jointly calibrated for dominant and secondary predictions using the same method. 

\end{abstract}

\section{Introduction}
Autonomous vehicles and other autonomous systems need to perceive their environment and detect and classify other traffic participants or objects in their vicinity. Based on these detections they plan their own behavior according to a high level goal. The planned behavior depends highly on the predicted classes and locations of the detected objects. Two attributes of object detectors are therefore important: It needs to precisely locate objects and classify them correctly, but it also must report meaningful information on the uncertainty of its predictions. If planning algorithms operate without precise uncertainty measures, they do not have the chance to adapt to the risk arising from the possibility of perceiving the environment incorrectly. Taking its own possible errors into account is a prerequisite for self-aware and explainable autonomous systems. In order to propagate uncertainties through all decision levels and incorporate them into planning, their initial estimates are of vital importance. We therefore aim to make the information about uncertainties arising in object detection more reliable. While there is uncertainty in both localization and classification, this research focuses on the classification uncertainty. 

Even though current object detectors do usually not output their localization uncertainty, there are different methods for estimating it, such as Monte Carlo Dropout \cite{pmlr-v48-gal16, NIPS2017_84ddfb34}, Deep Ensemble \cite{NIPS2017_9ef2ed4b}, Direct Modelling \cite{NIPS2017_2650d608, Gast_2018_CVPR} and Error Propagation \cite{153653e0ad994c6292e655ac8a382001}. However, not all of these approaches are feasible for real-world systems due to the computational overhead of Bayesian methods \cite{9525313}.

In contrast, a measure for the classification uncertainty is intrinsically available in all deep neural network based object detectors. They take an input $\bold{x}\in\mathcal{X}$ and output a confidence vector $\mathbf{p}\in[0,1]^C$ alongside the maximum likelihood estimates for the bounding box parameters. The confidence vector $\mathbf{p}$ can be interpreted as the discrete probability distribution over the $C$ classes from set $\mathcal{C}$. As of today, usually only the maximum element $p_i$ of $\bold{p}$ is regarded and determines $i$ as the object’s class if $p_i$ exceeds a certain threshold. This disregard of the actual confidence and the secondary predictions is sub-optimal for safety critical autonomous systems. Simple examples can be constructed to show that the optimal behavior changes if the dominant class is predicted with a different confidence level or the confidence ranking of the secondary classes change: An object is classified as a vehicle with a high confidence requires a different planning than an object with high classification ambiguity between inanimate object or vulnerable road user. To assess the risk of its actions properly, a planning algorithm needs to have access to correct confidence estimates not only for the dominantly predicted class, but also for the less confident secondary class predictions.

Guo et al. \cite{ChuanGuo.2017} show that the confidence estimates of modern neural network classifiers are usually poorly calibrated, which means that the predicted classification confidence deviates from the observed empirical value. Küppers et al. \cite{FabianKuppers.2020} extend their analysis to 2D object detectors coming to a similar result. Perfect calibration can be expressed in two ways: The strong calibration condition 
\begin{equation}\label{eq:strong_cal}
    \mathbb{P}(Y=i\mid p_i(\bold{x}))=p_i \text{, for all } i \in \mathcal{C} ,
\end{equation}
demands that for a set of detections the post-hoc measured empirical probability distribution of the object’s ground truth class equals the predicted distribution including the non-dominant secondary classes \cite{NEURIPS2019_1c336b80}.
The weak calibration condition  
\begin{equation}\label{eq:weak_cal}
    \mathbb{P}(Y=i \mid p_i(\bold{x})) = p_i \text{, for } i=\text{argmax }p_i(\bold{x})
\end{equation}
takes only the dominant class into account and allows a classifier to incorrectly rank secondary predictions while still regarding it as perfectly calibrated \cite{ChuanGuo.2017}.

While there is a large body of research on the calibration of classifiers starting together with the early developments of machine learning and often stemming from meteorological or medical backgrounds \cite{DeGroot.1983, DBLP:conf/nips/DenkerL90, Jiang.2012, 29d223435bcd45678a2f0f1d86b9ecd6}, the extent of the miscalibration of modern neural networks is still disputed and might depend strongly on the specific architecture \cite{ChuanGuo.2017, NEURIPS2021_8420d359}. Research about confidence calibration of object detectors is scarcer and mostly focused on 2D models operating on image data \cite{8417976, pathiraja2023multiclass, FabianKuppers.2020, 9575841, Küppers2022}. Confidence calibration of 3D object detectors is performed by \cite{DiFeng.2019} and \cite{Kato.642022692022}. However, their works only employ the post-hocs methods Isotonic Regression and Temperature Scaling. To the best of our knowledge, a broad comparison of different classification calibration methods on state-of-the-art 3D object detectors for autonomous driving is not available. 

\subsection*{Contribution}
Our contribution is threefold:
\begin{itemize}
    \item First, we deduce a new calibration metric called Full Detection Expected Calibration Error (Full D-ECE) from the strong calibration condition. In contrast to existing metrics, our metric therefore captures the calibration of the full predictive class distribution.
    \item Second, we introduce two different auxiliary loss terms $\mathcal{L}_{\text{DECE}}$ and $\mathcal{L}_{\text{FullDECE}}$, which induce regularization proportional to miscalibration during the training of object detectors and work as a train-time calibration method. $\mathcal{L}_{\text{FullDECE}}$ together with an adapted form of Isotonic Regression achieves best in class calibration for non transformer based 3D object detectors.
    \item Finally, we evaluate several combinations of train-time and post-hoc calibration methods on three different 3D object detectors on the Waymo Open dataset \cite{Sun_2020_CVPR} for autonomous driving.
\end{itemize}

\section{Related Work}
The calibration of predictors is a long standing topic of research. Post-hoc methods are developed and used since several decades while train-time methods attracted interest only recently with increased application of deep neural networks.
\subsection{Post-Hoc Calibration Methods}
Post-hoc calibration methods aim to calibrate the confidence estimates of classifiers or object detectors once their training is finished. They can be divided into binning and scaling methods. 

Binning methods achieve calibration by binning confidence values and remapping their estimates to the empirical accuracy or precision of the respective bin. 
Histogram Binning is a basic method which groups predictions according to their confidence levels, then assigns each detection the empirical accuracy calculated for the respective bin on the calibration data set \cite{B.Zadrozny.2001}.
Isotonic Regression extends Histogram Binning by optimizing both the confidence calibration within bins and the bin boundaries themselves \cite{B.Zadrozny.2001}.
Naeini et al. \cite{MahdiPakdamanNaeini.2015} introduce Bayesian Binning into Quantiles where they extend histogram binning by combining several binning models in a Bayesian probabilistic fashion.
An Ensemble of Near Isotonic Regression \cite{MahdiPakdamanNaeini.2015b} employs a similar strategy by finding a Bayesian combination of several models using Near Isotonic Regression \cite{R.Tibshirani.2011}, a variant of Isotonic Regression with relaxed monotonicity assumptions.
Binning methods are non-parametric, meaning that they do not employ underlying assumptions about the data distribution.

Scaling methods do not manipulate the confidence estimates directly but rather rescale the logits before they are passed to the softmax or sigmoid activation function of the output layer. 
For binary classification problems or class-agnostic calibration, Platt Scaling \cite{Platt2000} optimizes the parameters $a$ and $b$ of the function $y(x) = \sigma(a x +b)$, where $y$ is the class confidence, $x$ the logit before the activation function and $\sigma$ the sigmoid function. During training of the classifiers the parameters are set to $a=1$ and $b=0$. In the subsequent calibration step, the parameters are fitted using a hold out dataset while the weights of the classifier remain fixed.
Vector Scaling and Matrix Scaling \cite{ChuanGuo.2017} extend Platt Scaling for the class specific calibration of multi class classifiers. Instead of a scalar parameter, these methods use a vector or a matrix of learnable parameters to scale the logits for each class independently in case of Vector Scaling or dependently in case of Matrix Scaling.
Temperature Scaling \cite{ChuanGuo.2017} is a simpler variant of Platt Scaling, which uses a single same parameter to scale all class logits. Despite its simplicity, Temperature Scaling is effective in calibrating the class confidences while at the same time preserving the accuracy of the trained classifier \cite{JizeZhang.2020}. This makes Temperature Scaling a popular post-hoc calibration method. 

In contrast to binning methods, scaling methods as parametric methods apply assumptions about the distribution of their parameters. This allows the parametric methods to optimize their parameters using less training data. While Platt Scaling and its descendants assume a normal distribution, methodically similar approaches such as Beta \cite{MeelisKull.2017} and Dirichlet calibration \cite{Kull.2019} assume a Beta distribution for binary or a Dirichlet distribution for multi-class classification.
Zhang et al. \cite{JizeZhang.2020} show that combining parametric and non-parametric approaches can yield good calibration results.

\subsection{Train-Time Calibration Methods}
Label Smoothing, although designed to increase model performance and reduce overfitting, leads to inherently better calibrated models. It achieves it positive effect on calibration by removing the incentive for the model to place all probability mass on a single class while still enforcing separation in the logit values attributed to different classes \cite{NEURIPS2019_f1748d6b}, \cite{Pereyra.2017}.
Similarly, Focal Loss \cite{8417976} aims to mitigate the problem of hard and easy classes but also increases the entropy of the predictive confidence distribution. The resulting distribution is less pronounced and better calibrated \cite{Mukhoti.2020}.
Adaptive Focal Loss (AdaFocal) \cite{NEURIPS2022_0a692a24} is an extension of focal loss specifically designed for model calibration. During training, the model’s miscalibration is constantly observed and the Focal Loss parameter $\gamma$ is adapted to increase or decrease the gradients and thereby counteracting under-confident or overconfident predictions. 

Other researchers introduce auxiliary loss terms which can be added as a regularization to any loss function. The Multiclass Difference of Confidence and Accuracy (MDCA) calculates the average confidence and accuracy on the current mini-batch and adds their difference as a regularizing component to the loss \cite{9879475}.
The auxiliary Multiclass Confidence and Localization Calibration loss \cite{pathiraja2023multiclass} introduces two additional loss terms based on classification confidence uncertainty and localization uncertainty. During training, Monte-Carlo dropout is used to estimate the uncertainties of confidence predictions and bounding box parameters. These estimates are compared to the predicted confidence and the overlap with the ground-truth bounding box.
\section{Methods}
\subsection{Existing Metrics for Calibration}
The most widespread metric to measure the miscalibration of classifiers is the Expected Calibration Error (ECE) \cite{MahdiPakdamanNaeini.2015}. The ECE evaluates the expected divergence from the weak calibration condition (\ref{eq:weak_cal}). It approximates
\begin{equation}\label{eq:exp_ECE}
    \mathbb{E}(\mathbb{P}(Y=i \mid p_i(\bold{x})) - p_i)\text{, for } i=\text{argmax }p_i(\bold{x})
\end{equation}
by dividing the confidence predictions into $B$ equally-spaced bins $\mathcal{B}$, with $n_b$ out of total $N$ predictions falling into a bin $b$, and comparing the difference between each bin's accuracy (fraction of correct predictions) and average confidence (mean predicted probability). The ECE is then computed as the weighted average of these calibration errors across all bins,
\begin{equation}\label{eq:ECE}
    \text{ECE} = \sum_{b=1}^B \frac{n_b}{N} \cdot \lvert \text{acc}(b) - \text{conf}(b) \rvert \text{.}
\end{equation}
Intuitively, a classifier is perfectly calibrated if the mean predicted confidence equals the empirical probability of correct predictions for all confidence bins.

In object detection, True Negatives and therefore accuracy are not defined. Küppers et al. \cite{FabianKuppers.2020} introduce a metric which uses precision instead of accuracy while also refining the binning scheme. Their metric enables binning based on bounding box parameters (such as position or size) in addition to confidence bins. This allows better evaluation of location-dependent miscalibration. However, binning across these additional bounding box dimensions requires an exponentially larger number of predictions to calculate meaningful precision and averaged confidence values. They define the Detection Expected Calibration Error (D-ECE) as

\begin{equation}
    \text{D-ECE} = \sum_{b=1}^{B_{\text{total}}} \frac{n_b}{N} \cdot \lvert \text{prec}(b) - \text{conf}(b) \rvert,
\end{equation}

where $B_{\text{total}} = \prod_{k=1}^{K} B_k$, obtained by iterating over all $K$ dimensions used for binning.

\subsection{Full D-ECE Calibration Metric}
Both ECE and D-ECE are based on the weak condition for confidence calibration (\ref{eq:weak_cal}) and take only the predicted class into account. This is a substantial shortcoming when the optimal action of an autonomous system also depends on secondary class predictions, as in the example of an ambiguous detection with vulnerable road users involved. To be able to measure the calibration regarding the distribution of predicted probabilities for all classes, we introduce the Full Detection Expected Calibration Error (Full D-ECE). In analogy to (\ref{eq:exp_ECE}), it is derived from the strong calibration condition (\ref{eq:strong_cal}) to estimate
\begin{equation}
    \mathbb{E}(\mathbb{P}(Y=i \mid p_i(\bold{x})) - p_i)\text{, for all } i \in \mathcal{C}.
\end{equation}
To calculate the Full D-ECE, we follow the same logic as the ECE and D-ECE only that we take the full vector of predicted probabilities for all $C$ classes. This means that each detected object generates one “true positive” confidence prediction $p_{c,\text{ true}}$ and $C-1$  “false positive” confidence predictions. We replace the precision of bin $b$ with the ratio of the number of confidence predictions for correct classes falling into confidence bin $b$ (regardless whether the ground truth class is assigned the highest probability or not) and total predictions in $b$, $\frac{\lvert \mathcal{Y}_{\text{tp, b}} \rvert}{n_b}$. The average bin confidence $\text{conf}(b)$ is now calculated over all confidence estimates for correct as well as wrong classes in bin $b$ and therefore becomes $\frac{\sum_{p \in \mathcal{Y}_b}p}{n_b}$. More precisely and after reduction of $n_b$,
\begin{equation}\label{eq:fulldece}
    \text{Full D-ECE} = \sum_{b=1}^{B_{\text{total}}} \frac{\lvert \lvert \mathcal{Y}_{\text{tp, b}} \rvert - \sum_{p \in \mathcal{Y}_b}p \rvert}{N},
\end{equation}
with $p$ being the elements of the $C$-dimensional confidence vector for one of the $N$ total detections and sets $\mathcal{Y}_b=\{p_{c,n}|c\in \mathcal{C}, n \in\mathcal{N}, p\in\mathcal{B}_b\}$ and $\mathcal{Y}_{\text{tp, b}}=\{p_{c,n}| c=c_{\text{true}}, n\in\mathcal{N}, p\in\mathcal{B}_b\}$, where $\mathcal{C}$ and $\mathcal{N}$ denote the sets of all classes and detections.

The Full D-ECE metric comes with another advantage: While the D-ECE metric is calculated only from the confidence of the predicted class, low confidence values are underrepresented, making the metric imprecise where the classifier is already relatively uncertain. The Full D-ECE on the other hand increases the number of available confidence values and has a fixed ratio of "true positive" and "false positive" confidence predictions. This results in more densely populated low-confidence bins, allowing more accurate statements about calibration in this range.  

\subsection{D-ECE and Full D-ECE as Auxiliary Loss Functions}
We propose to enforce confidence calibration by including the confidence metrics D-ECE and Full D-ECE as auxiliary loss functions in addition to the base classification loss
\begin{equation}
    \mathcal{L}_{\text{class}}= \mathcal{L}_{\text{base}}+\alpha\mathcal{L}_{\text{aux}},
\end{equation}
with $\alpha$ as the weight determining hyperparamter of the auxiliary loss term and $\mathcal{L}_{\text{aux}} = \mathcal{L}_{\text{DECE}} = \text{D-ECE}$ or $\mathcal{L}_{\text{aux}} = \mathcal{L}_{\text{FullDECE}} = \text{Full D-ECE}$.

This idea is not new since \cite{9879475} and \cite{29d223435bcd45678a2f0f1d86b9ecd6} also introduce a regularizing term based on the error between average confidence and empirical accuracy (or precision for object detection). However, both works do not use confidence binning to calculate the deviation from perfect calibration. This means that overconfident and underconfident predictions can cancel each other out, resulting in strong miscalibration as measured by the confidence-binned calibration metrics, but weak corresponding regularization in the loss function. In contrast, our approach utilizes the confidence-binned deviation from perfect calibration and regularizes the network using the D-ECE or Full D-ECE directly as auxiliary loss term weighted with a hyperparameter $\alpha$. Both \cite{9879475} and \cite{29d223435bcd45678a2f0f1d86b9ecd6} state that confidence binning leads to a non-differentiable loss function, which is true since it has non-continuities at the bin borders. However, in practice this is not a problem and there are many common loss functions, such as ReLU, which are not differentiable. Because the probability of a confidence value landing exactly on the bin boundary converges to zero in the continuous case and is vanishingly small in the quantized case, we can use the standard ruleset of the PyTorch library for dealing with locally not-differentiable functions \cite{.1262025}.
\section{Experiments}
For our analysis, we select three LiDAR point cloud detectors based on their real-world applicability to autonomous vehicles and their architectural diversity. CenterPoint \cite{9578166}, a former state-of-the-art two-stage detector, remains widely used as the default detector in Autoware's open-source autonomous driving stack. PillarNet \cite{10.1007/978-3-031-20080-9_3                                                                 } offers real-time detection in a single-stage architecture. DSVT \cite{10203294} employs dynamic sparse window attention for point cloud feature encoding. We evaluate these detectors' miscalibration and compare both post-hoc and train-time calibration methods to assess our auxiliary calibration losses and identify the most effective confidence calibration strategy.

\subsection{Experimental Setup}
We use the OpenPCDet framework \cite{openpcdet2020} to train CenterPoint and PillarNet for 60 epochs on the Waymo Open Dataset's \cite{Sun_2020_CVPR} training split downsampled to 2 Hz, while DSVT is trained for 24 epochs on the original training dataset sampled with 10 Hz. Each training is conducted with the largest feasible batch size of 45. For each network, we train a baseline using the default training regimen with voxel based focal loss for classification. Additionally, we train each model using either a combination of the standard loss and Adaptive Focal Loss with $\gamma$ according to \cite{NEURIPS2022_0a692a24} or add D-ECE or Full D-ECE auxiliary loss terms to the focal loss. The calibration evaluation necessary during training uses 15 confidence bins and is performed on the last three batches for the Adaptive Focal Loss and on the last two batches for $\mathcal{L}_{\text{DECE}}$ and $\mathcal{L}_{\text{FullDECE}}$. We explore hyperparameter values $\alpha = [0.5, 1, 2, 5, 10, 20]$ for our auxiliary losses and find $\alpha =1$ to produce the best results.  

We further calibrate all models with the post-hoc methods Temperature Scaling, Platt Scaling, and Isotonic Regression using the implementation by \cite{10.1007/978-3-031-72664-4_11                           }. For this purpose, we split the test set into two equal parts for calibration and evaluation. We fit the parameters for Temperature Scaling and Platt Scaling only on the dominant predictions (the classical approach aiming for the weak calibration condition (\ref{eq:weak_cal})). Since Isotonic Regression has more degrees of freedom, we fit it not only on the dominant prediction but also incorporate the secondary predictions. Post-hoc calibration and evaluation of miscalibration uses 25 confidence bins.

\subsection{Miscalibration of 3D Object Detectors}
Tab. \ref{tab:baseline} provides a comparison of calibration performance for the baseline object detectors. In agreement with \cite{ChuanGuo.2017}, we find that the more complex DSVT-Pillar network is less calibrated than the simpler CenterPoint and PillarNet networks ass measured by D-ECE. Nevertheless, it achieves a slightly better Full D-ECE, indicating improved calibration for secondary predictions. We note however that small confidence estimates are overrepresented in the Full D-ECE score because secondary predictions tend to be less confident while outweighing the dominant predictions by numbers if there are more than two classes.

\begin{table}[pbth]
\caption{Comparison of baseline object detectors}
    \begin{center}
    \begin{tabular}{l|c|c|c|c}
         Network & mAP (L1) & mAP (L2) & D-ECE & Full D-ECE  \\
         \hline
         CenterPoint \cite{9578166} & 0.734 & 0.673 & 0.159 & 0.075\\
         PillarNet \cite{10.1007/978-3-031-20080-9_3                                                                    } & 0.722 & 0.661 & 0.156 & 0.076\\
         DSVT \cite{10203294} & 0.795 & 0.732 & 0.286 & 0.070\\
    \end{tabular}
    \end{center}
    \label{tab:baseline}
\end{table}

\begin{figure*}[htbp]
\centering
\begin{subfigure}[t]{0.45\textwidth}
        \centering
        \fontsize{8pt}{12pt}\selectfont
        \def\svgwidth{0.98\linewidth}
        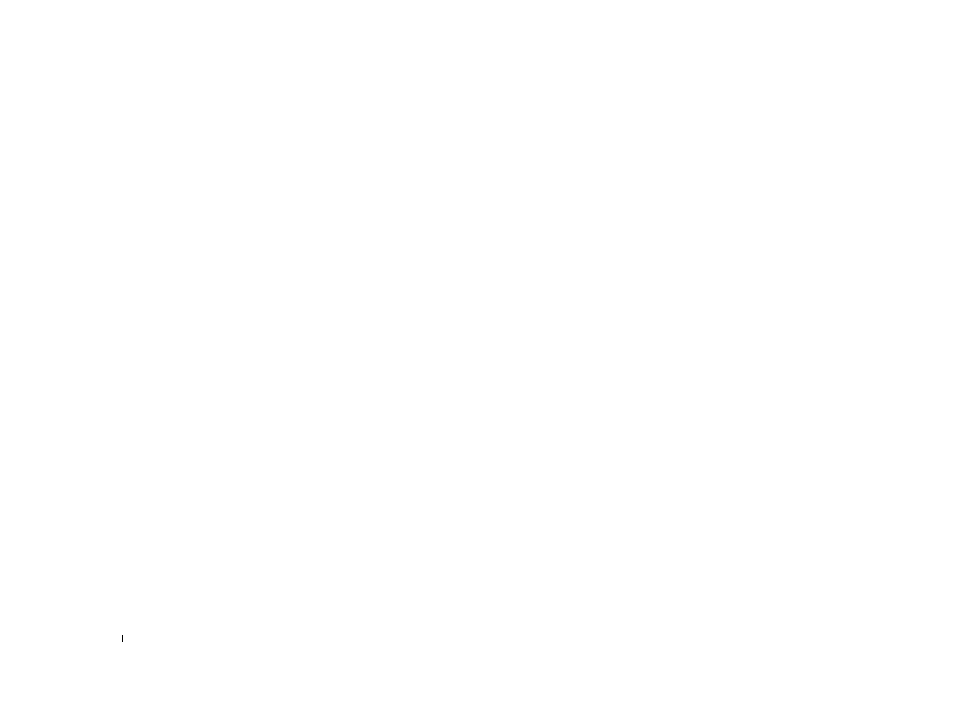
        \caption{Dominant predictions}
\end{subfigure}
\begin{subfigure}[t]{0.45\textwidth}
        \centering
        \fontsize{8pt}{12pt}\selectfont
        \def\svgwidth{0.98\linewidth}
        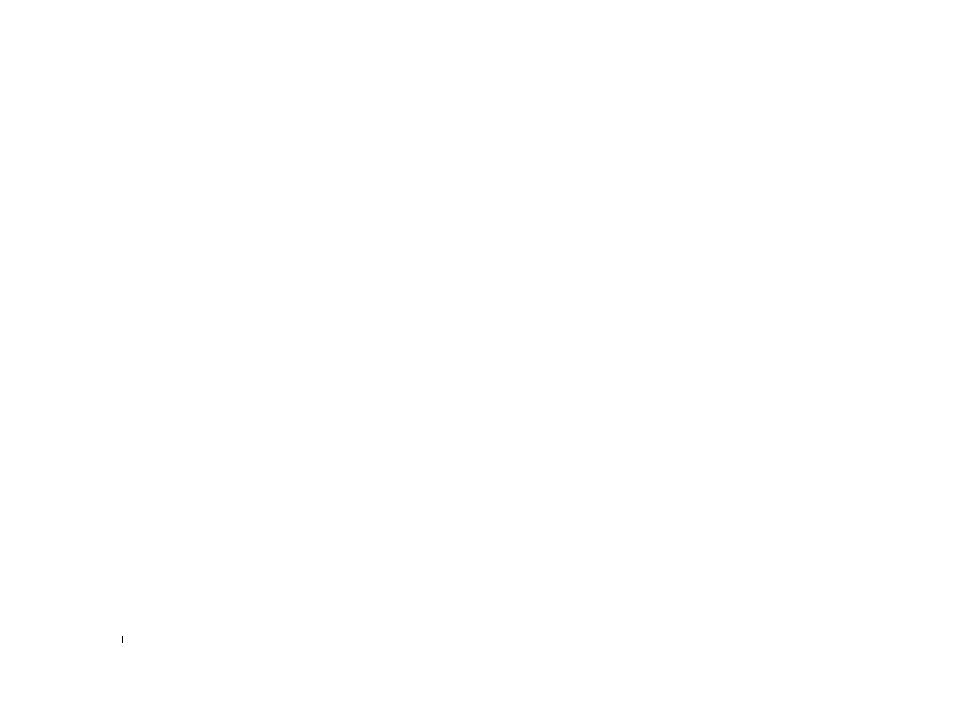
        \caption{Dominant and secondary predictions}
\end{subfigure}
\caption{When evaluating the miscalibration of the baseline detectors for dominant predictions only (a), CenterPoint and PointPillar are better calibrated than DSVT-Pillar. If the full vector of confidence predictions is regarded (b), DSVT-Pillar becomes less overconfident for low to mid confidence levels while the other detectors increase their overconfidence.}
\label{fig:calib_bl}
\end{figure*}

Fig. \ref{fig:calib_bl} shows the miscalibration of the baseline object detectors over the range of predicted confidence values taking into account either the dominant class prediction or the full confidence vector. We note that for DSVT-Pillar overconfidence is reduced for low to mid confidence levels when regarding calibration according to the strong condition (\ref{eq:strong_cal}). In contrast, CenterPoint and Pillarnet increase their overconfidence when calibration is evaluated on dominant and secondary predictions.

\subsection{Confidence Calibration}
Tab. \ref{tab:calib} summarizes the results of applying train-time and post-hoc calibration methods individually and in combination to the three baseline object detectors. Notably, each network achieves its highest mAP and mAPH scores either in its uncalibrated baseline form or when calibrated with Temperature or Platt Scaling.

\begin{table*}[hbtp]
\caption{Calibration results: Train-time and posth-hoc methods. \textbf{Bold} marks the best results for each object detector.}
    \begin{center}
    \begin{tabular}{l l|c|c|c|c}
         Network & Calibration Method & mAP / mAPH (L1) $\uparrow$ & mAP / mAPH (L2) $\uparrow$ &  D-ECE $\downarrow$ & Full D-ECE $\downarrow$  \\
         \hline
         CenterPoint \cite{9578166} & uncalibrated & \textbf{0.734 / 0.708} & \textbf{0.673 / 0.649} & 0.159 & 0.075\\ 
         & Temperature Scaling & 0.733 / 0.707 & 0.673 / 0.648 & 0.145 & 0.056\\
         & Platt Scaling & \textbf{0.734 / 0.708} & \textbf{0.673 / 0.649} & 0.080 & 0.032\\
         & Isotonic Regression & 0.696 / 0.666 & 0.640 / 0.611 & 0.078 & 0.106\\
         & Isotonic Regression, full & 0.693 / 0.664 & 0.633 / 0.606 & 0.031 & \textbf{0.013}\\
          & AdaFocal & 0.676 / 0.649 & 0.612 / 0.588 & $\uparrow$ 0.355 $\text{  }$& 0.135 \\
          & AdaFocal + Temperature Scaling & 0.658 / 0.628 & 0.595 / 0.569 & 0.352 & $\uparrow$ 0.450 $\text{  }$\\
          & AdaFocal + Platt Scaling & 0.666 / 0.639 & 0.603 / 0.578 & 0.082 & 0.020\\
          & AdaFocal + Isotonic Regression &  $\downarrow$ 0.618 / 0.587 $\text{  }$ & $\downarrow$ 0.558 / 0.530 $\text{  }$& 0.063 & 0.085\\
          & AdaFocal + Isotonic Regression, full & 0.672 / 0.644 & 0.607 / 0.582 & 0.033 & 0.072\\
          & $\mathcal{L}_{\text{DECE}}$ & 0.731 / 0.704 & 0.671 / 0.646 & 0.113 & 0.069\\
          & $\mathcal{L}_{\text{DECE}}$ + Temperature Scaling & 0.732 / 0.704 & 0.671 / 0.646 & 0.102 & 0.052\\
          & $\mathcal{L}_{\text{DECE}}$ + Platt Scaling & 0.732 / 0.705 & 0.671 / 0.646 & 0.078 & 0.044\\
          & $\mathcal{L}_{\text{DECE}}$ + Isotonic Regression & 0.690 / 0.658 & 0.633 / 0.604 & 0.075 & 0.109\\
          & $\mathcal{L}_{\text{DECE}}$ + Isotonic Regression, full & 0.706 / 0.675 & 0.648 / 0.619 & 0.040 & 0.017 \\
          & $\mathcal{L}_{\text{FullDECE}}$ & 0.731 / 0.705 & 0.669 / 0.645 & 0.167 & 0.080\\
          & $\mathcal{L}_{\text{FullDECE}}$ + Temperature Scaling & 0.729 / 0.703 & 0.668 / 0.644 & 0.139 & 0.054\\
          & $\mathcal{L}_{\text{FullDECE}}$ + Platt Scaling & 0.730 / 0.705 & 0.669 / 0.645 & 0.079 & 0.034\\
          & $\mathcal{L}_{\text{FullDECE}}$ + Isotonic Regression & 0.698 / 0.668 & 0.639 / 0.611 & 0.074 & 0.103\\
          & $\mathcal{L}_{\text{FullDECE}}$ + Isotonic Regression, full & 0.707 / 0.678 & 0.647 / 0.620 & \textbf{0.030} & 0.014\\
          \cline{1-6}
        
         PillarNet \cite{10.1007/978-3-031-20080-9_3                                         } & uncalibrated & \textbf{0.722 / 0.687} & \textbf{0.661 / 0.629} & 0.156 & 0.076\\
         & Temperature Scaling & 0.721 / 0.687 & \textbf{0.661 / 0.629} & 0.146 & 0.059 \\
         & Platt Scaling & \textbf{0.722 / 0.687} & \textbf{0.661 / 0.629} & 0.081 & 0.031\\
         & Isotonic Regression & 0.688 / 0.649 & 0.631 / 0.595 & 0.073 & 0.103\\
         & Isotonic Regression, full & 0.694 / 0.655 & 0.637 / 0.601 & 0.031 & \textbf{0.013}\\
          & AdaFocal & 0.627 / 0.593 & 0.565 / 0.535 & $\uparrow$ 0.376 $\text{  }$ & $\uparrow$ 0.143 $\text{  }$\\
          & AdaFocal + Temperature Scaling & 0.574 / 0.535 & 0.518 / 0.483 & 0.262 & 0.085\\
          & AdaFocal + Platt Scaling & 0.627 / 0.593 & 0.565 / 0.535 & 0.062 & \textbf{0.013}\\
          & AdaFocal + Isotonic Regression & 0.586 / 0.549 & 0.529 / 0.496 & 0.056 & 0.082\\
          & AdaFocal + Isotonic Regression, full & 0.618 / 0.583 & 0.557 / 0.525 & 0.038 & 0.068\\
          & $\mathcal{L}_{\text{DECE}}$ & 0.719 / 0.685 & 0.658 / 0.627 & 0.164 & 0.081\\
          & $\mathcal{L}_{\text{DECE}}$ + Temperature Scaling & 0.718 / 0.684 & 0.657 / 0.626 & 0.135 & 0.056\\
          & $\mathcal{L}_{\text{DECE}}$ + Platt Scaling & 0.719 / 0.685 & 0.658 / 0.627 & 0.077 & 0.035\\
          & $\mathcal{L}_{\text{DECE}}$ + Isotonic Regression & 0.685 / 0.646 & 0.627 / 0.592 & 0.076 & 0.107\\
          & $\mathcal{L}_{\text{DECE}}$ + Isotonic Regression, full & 0.695 / 0.657 & 0.636 / 0.602 & \textbf{0.029} & 0.014\\
          & $\mathcal{L}_{\text{FullDECE}}$ & 0.718 / 0.684 & 0.657 / 0.626 & 0.169 & 0.079\\
          & $\mathcal{L}_{\text{FullDECE}}$ + Temperature Scaling & 0.717 / 0.682 & 0.656 / 0.624 & 0.144 & 0.056\\
          & $\mathcal{L}_{\text{FullDECE}}$ + Platt Scaling & 0.718 / 0.684 & 0.657 / 0.626 & 0.080 & 0.031\\
          & $\mathcal{L}_{\text{FullDECE}}$ + Isotonic Regression & 0.685 / 0.646 & 0.628 / 0.592 & 0.070 & 0.102\\
          & $\mathcal{L}_{\text{FullDECE}}$ + Isotonic Regression, full & 0.693 / 0.655 & 0.634 / 0.599 & \textbf{0.029} & \textbf{0.013}\\
          \cline{1-6}
         DSVT-Pillar \cite{10203294} & uncalibrated & \textbf{0.795 / 0.770} & \textbf{0.732 / 0.709} & 0.286 & 0.070\\
         & Temperature Scaling & \textbf{0.795 / 0.770} & \textbf{0.732 / 0.709} & 0.291 & 0.082\\
         & Platt Scaling & \textbf{0.795 / 0.770} & \textbf{0.732 / 0.709} & 0.076 & 0.064\\
         & Isotonic Regression & 0.773 / 0.744 & 0.713 / 0.686 & 0.047 & 0.092 \\
         & Isotonic Regression, full & 0.770 / 0.746 & 0.708 / 0.685 & 0.233 & 0.025\\
          & AdaFocal & 0.792 / 0.768 & 0.730 / 0.707 & $\uparrow$ 0.324 $\text{  }$  & 0.088\\
          & AdaFocal + Temperature Scaling & 0.792 / 0.767 & 0.729 / 0.706 & 0.306 & $\uparrow$ 0.115 $\text{  }$ \\
          & AdaFocal + Platt Scaling & 0.793 / 0.769 & 0.730 / 0.707 & 0.079 & 0.054\\
          & AdaFocal + Isotonic Regression & 0.767 / 0.739 & 0.707 / 0.681 & \textbf{0.044} & 0.080\\
          & AdaFocal + Isotonic Regression, full & 0.788 / 0.763 & 0.725 / 0.702 & 0.213 & \textbf{0.021}\\
          & $\mathcal{L}_{\text{DECE}}$ & 0.793 / 0.768 & 0.731 / 0.707 & 0.294 & 0.074\\
          & $\mathcal{L}_{\text{DECE}}$ + Temperature Scaling & 0.793 / 0.768 & 0.731 / 0.707 & 0.292 & 0.090\\
          & $\mathcal{L}_{\text{DECE}}$ + Platt Scaling & 0.794 / 0.769 & 0.731 / 0.707 & 0.078 & 0.064\\
          & $\mathcal{L}_{\text{DECE}}$ + Isotonic Regression & $\downarrow$ 0.766 / 0.736 $\text{  }$ & $\downarrow$ 0.707 / 0.679 $\text{  }$ & 0.046 & 0.094\\
          & $\mathcal{L}_{\text{DECE}}$ + Isotonic Regression, full & 0.777 / 0.752 & 0.709 / 0.686 & 0.234 & 0.026\\
          & $\mathcal{L}_{\text{FullDECE}}$ & 0.792 / 0.766 & 0.730 / 0.706 & 0.294 & 0.073\\
          & $\mathcal{L}_{\text{FullDECE}}$ + Temperature Scaling & 0.792 / 0.766 & 0.730 / 0.706 & 0.295 & 0.088\\
          & $\mathcal{L}_{\text{FullDECE}}$ + Platt Scaling & 0.792 / 0.767 & 0.730 / 0.706 & 0.079 & 0.064\\
          & $\mathcal{L}_{\text{FullDECE}}$ + Isotonic Regression & 0.768 / 0.738 & 0.708 / 0.680 & 0.045 & 0.093\\
          & $\mathcal{L}_{\text{FullDECE}}$ + Isotonic Regression, full & 0.774 / 0.748 & 0.712 / 0.687 & 0.238 & 0.026\\
    \end{tabular}
    \end{center}
    \label{tab:calib}
\end{table*}

Platt Scaling consistently improves calibration, as measured by both D-ECE and Full D-ECE, for all three baseline detectors and their variants with train-time calibration while preserving detection quality.

Isotonic Regression applied only to the dominant predictions, following \cite{DiFeng.2019} and \cite{Kato.642022692022}, effectively reduces D-ECE across all model and train-time configurations. However, it slightly increases Full D-ECE and significantly reduces mAP and mAPH. When Isotonic Regression is instead fitted to the full prediction vectors, the Full D-ECE score decreases substantially in all networks, regardless of whether they employ an auxiliary train-time method. For CenterPoint and PillarNet, the D-ECE also decreases. In contrast, DSVT-Pillar is not effectively calibrated in terms of D-ECE by this approach. This is due to the fact that the discrepancy in calibration between dominant predictions and the full confidence vector is larger than for the other detectors. As depicted in fig. \ref{fig:calib_bl}, the full confidence vectors of DSVT suffer from less miscalibration than the dominant predictions alone. Therefore, using secondary predictions for isotonic regression does not scale the dominant predictions sufficiently, especially in the low confidence range where the number of secondary predictions outweighs dominant predictions. 

Nevertheless, fitting Isotonic Regression to non-dominant predictions yields higher mAP for all networks and training configurations compared to using dominant predictions alone. The other post-hoc methods do not benefit from fitting on secondary predictions. Since they are restricted to a specific functional form, they easily overfit to the many low confidence secondary predictions. Isotonic Regression on the other hand, is only restricted to a monotonically increasing function with individual mapping for each of its bins. Therefore, it calibrates the fewer, usually dominant, predictions in the high-confidence bins unaffected by the large amount of low-confidence predictions.

In experiments with Adaptive Focal Loss, CenterPoint and PillarNet experience a pronounced drop in mAP and mAPH relative to baseline focal loss and our auxiliary train-time losses. This effect is not observed in DSVT-Pillar. The calibration results for Adaptive Focal Loss vary widely: as a stand-alone method, it worsens miscalibration in all detectors. However, when combined with Platt Scaling, it achieves decent calibration in D-ECE and Full D-ECE scores for each network.

Our auxiliary training losses, $\mathcal{L}_{\text{DECE}}$ and $\mathcal{L}_{\text{FullDECE}}$, maintain detection quality on par with the baseline networks. $\mathcal{L}_{\text{DECE}}$ improves the calibration of CenterPoint relative to its uncalibrated baseline, but slightly degrades calibration in PillarNet and DSVT. Meanwhile, $\mathcal{L}_{\text{FullDECE}}$ is ineffective as a stand-alone calibration method; in combination with Isotonic Regression (standard or full), however, it achieves best in class calibration for D-ECE on CenterPoint and for both metrics on PillarNet.

\begin{figure*}[pbth]
\centering
\begin{subfigure}[t]{0.3\textwidth}
        \centering
        \fontsize{8pt}{12pt}\selectfont
        \def\svgwidth{0.98\linewidth}
        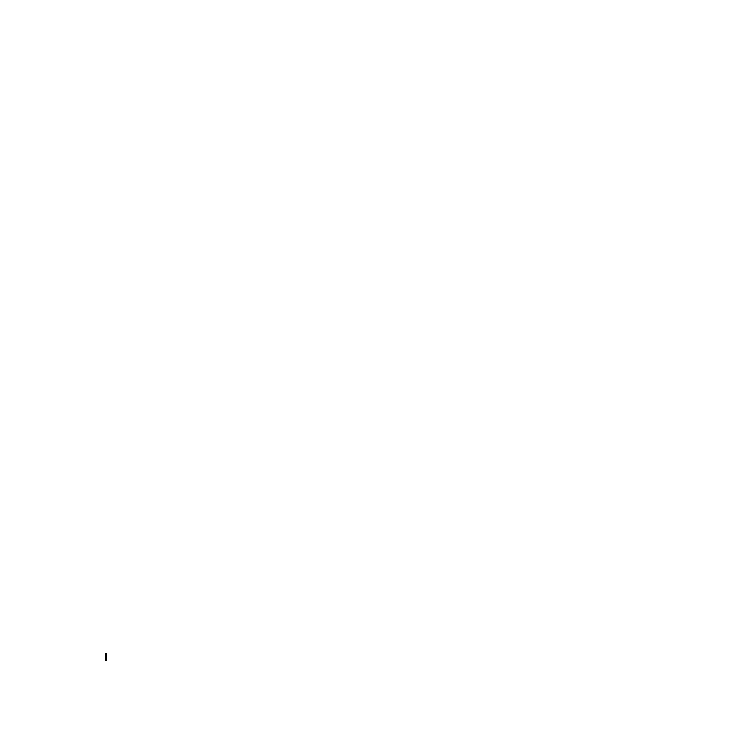
        \caption{CenterPoint, $\mathcal{L}_{\text{FullDECE}}$ + Isotonic Regression (full)}
\end{subfigure}
\begin{subfigure}[t]{0.3\textwidth}
        \centering
        \fontsize{8pt}{12pt}\selectfont
        \def\svgwidth{0.98\linewidth}
        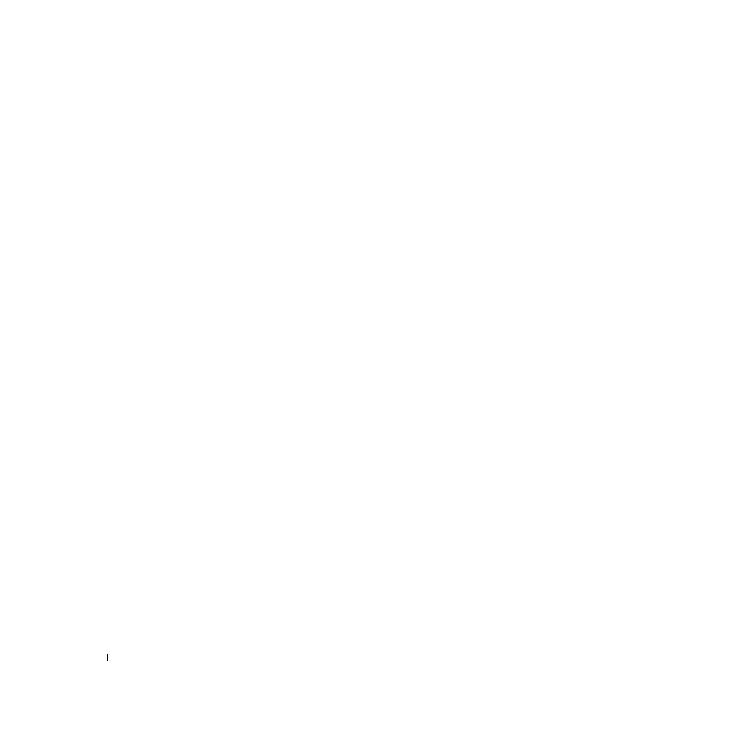
        \caption{PillarNet, $\mathcal{L}_{\text{FullDECE}}$ + Isotonic Regression (full)}
\end{subfigure}
\begin{subfigure}[t]{0.3\textwidth}
        \centering
        \fontsize{8pt}{12pt}\selectfont
        \def\svgwidth{0.98\linewidth}
        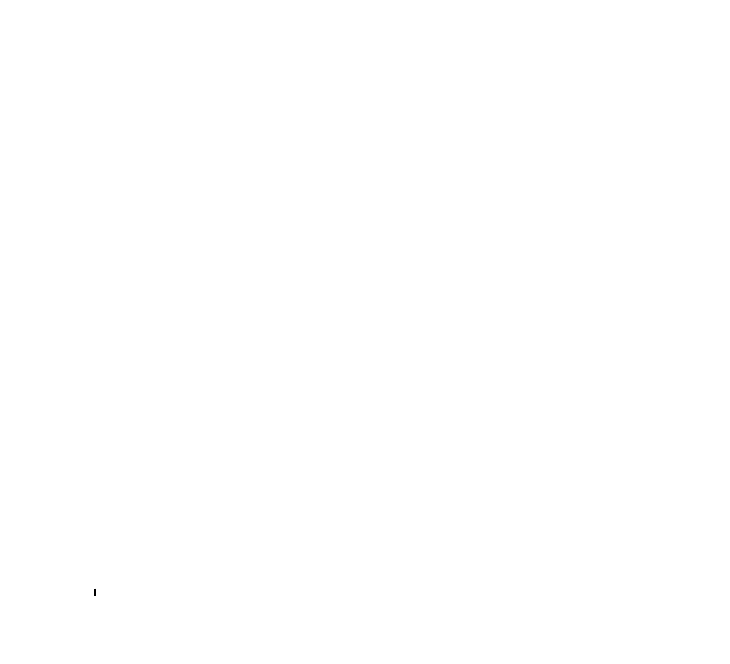
        \caption{DSVT-Pillar, Adaptive Focal Loss + Isotonic Regression (dominant/full)}
\end{subfigure}
\caption{We are able to simultaneously calibrate CenterPoint (a) and PillarNet (b) effectively for dominant predictions only and the full confidence vector using a combination of our training loss $\mathcal{L}_{\text{FullDECE}}$ and Isotonic Regression fitted on all predictions. Because simultaneous calibration of dominant predictions and full confidence vector does not succeed in case of DSVT-Pillar (c), we fit Isotonic Regression for calibration of dominant predictions and the full confidence vector only on dominant or all predictions.}
\label{fig:results}
\end{figure*}

We achieve the best calibration as measured by D-ECE for CenterPoint and PillarNet with the combination of $\mathcal{L}_{\text{FullDECE}}$ and Isotonic Regression fitted on the full confidence vector (fig. ref{fig:results}). Interestingly, we are able to calibrate the dominant and secondary predictions simultaneously. Although the Full D-ECE scores for both networks are smaller than the D-ECE, fig. \ref{fig:results}(a) and \ref{fig:results}(b) suggest better calibration of the dominant class prediction. Further analysis shows, that the Full D-ECE metric is heavily influenced by the large amount of samples in the lowest confidence bin, where miscalibration is low.

In case of DSVT, we achieve effective calibration for both dominant and secondary confidence predictions using Adaptive Focal Loss and Isotonic Regression, as depicted in fig. \ref{fig:results}(c). However, no combination of calibration methods was able to do so simultaneously. The decision on which confidence predictions to fit Isotonic Regression influences the calibration of dominant and secondary predictions differently since dominant predictions and full confidence vectors exhibit a different miscalibration pattern. It is therefore beneficial to choose the post-hoc method according to whether only dominant or all confidence predictions will be used in an application.

\section{Conclusion}
From our experiments we find that it is indeed possible to achieve good calibration of 3D object detectors not only for dominant but also for secondary class predictions. This distinction is especially important in safety critical applications. However, no single strategy works equally well for all detectors. If retaining the original detection quality is paramount, Platt Scaling is a good choice; either as stand-alone method or in combination with $\mathcal{L}_{\text{DECE}}$ or $\mathcal{L}_{\text{FullDECE}}$ auxiliary losses for Centerpoint and Pillarnet or combined with Adaptive Focal Loss for the more complicated DSVT-Pillar detector. For best-in-class calibration, CenterPoint and PillarNet benefit most from combining our regularizing loss term $\mathcal{L}_{\text{FullDECE}}$ with Isotonic Regression fitted on the entire confidence vector. The DSVT-Pillar transformer-based detector, however, reveals a different miscalibration pattern which makes it challenging to optimize both dominant and secondary predictions simultaneously. The best combination is Adaptive Focal Loss with Isotonic Regression; but it is important to decide before calibration whether the focus lies on dominant predictions or the full confidence vector. Future research should thus emphasize calibration for transformer-based networks. We furthermore encourage to evaluate next-generation 3D detectors not only on detection metrics but also on calibration quality and adaptability to various calibration strategies. 

\section*{Acknowledgment}
Author contributions: C. Schröder, as the first author,
devised the essential concepts of the proposed metric and auxiliary loss terms, conducted the analysis and wrote the article.
M.-R. Schlüter implemented the concepts in code.
M. Lienkamp made an essential contribution to the concept of the research project. He revised the paper critically for important intellectual content. M. Lienkamp gives final approval for the version to be published and agrees to all aspects of the work. As a guarantor, he accepts responsibility for the overall integrity of the paper.

\bibliographystyle{IEEEtran}
\bibliography{Calibration_File}
\end{document}